\documentclass[twocolumn,letterpaper]{fairmeta}

\usepackage{geometry}

\usepackage{xspace}
\RequirePackage{amsmath}
\RequirePackage{amssymb}
\makeatletter
\DeclareRobustCommand\onedot{\futurelet\@let@token\@onedot}
\def\@onedot{\ifx\@let@token.\else.\null\fi\xspace}

\usepackage{balance}

\usepackage{algorithm}
\usepackage{algorithmic}
\usepackage{listings}
\usepackage[most]{tcolorbox}
\newtcolorbox{prompt}[1]{
    left=4mm,
    right=4mm,
    top=2mm,
    bottom=2mm,
    boxsep=0mm,
    rounded corners,
    title=#1,
    fontupper=\footnotesize\linespread{0.9}\fontfamily{lmr}\selectfont,
    }

\makeatother

\definecolor{classicrose}{rgb}{0.98, 0.8, 0.91}
\definecolor{flavescent}{rgb}{0.97, 0.91, 0.56}
\definecolor{grannysmithapple}{rgb}{0.66, 0.89, 0.63}
\definecolor{languidlavender}{rgb}{0.84, 0.79, 0.87}
\definecolor{lightblue}{rgb}{0.68, 0.85, 0.9}

\newcommand{\ourtool}{\textit{Source2Synth}\xspace}
\newcommand{\FTone}{\textit{LLMSynth}\xspace}
\newcommand{\FTonecomb}{\textit{LLMSynth-datamix}\xspace}
\newcommand{\FTtwo}{\textit{LLMCurated}\xspace}
\newcommand{\FTtwocomb}{\textit{LLMCurated-datamix}\xspace}

\usepackage{amsfonts}

\title{Source2Synth: Synthetic Data Generation and Curation \\Grounded in Real Data Sources}

\author[1,2]{Alisia Lupidi}
\author[1,\dagger]{Carlos Gemmell}
\author[1]{Nicola Cancedda}
\author[1]{Jane Dwivedi-Yu}
\author[1]{Jason Weston}
\author[1,2]{Jakob Foerster}
\author[1]{Roberta Raileanu}
\author[1]{Maria Lomeli}

\affiliation[1]{Meta Superintelligence labs}
\affiliation[2]{University of Oxford}

\contribution[\dagger]{Work done during internship at Meta}

\abstract{Synthetic data generation has recently emerged as a promising approach for enhancing the capabilities of large language models (LLMs) without the need for expensive human annotations. However, existing methods often generate data that can be low quality or contrived. In this paper, we introduce~\ourtool{}, a scalable approach for synthetic data generation and curation that is grounded in real-world data sources.~\ourtool{} takes as input a custom data source and produces synthetic data examples with intermediate reasoning steps. Our method improves the dataset quality by discarding low-quality generations based on their answerability. We demonstrate the generality of this approach by applying it to two tasks that leverage two different types of data: multi-hop question answering (MHQA), where we test complex reasoning abilities leveraging documents, and tabular question answering (TQA), where we test tool usage leveraging tables. Our method improves performance by 25.51\% for TQA on WikiSQL and 22.57\% for MHQA on HotpotQA compared to the fine-tuned baselines.}
\date{\today}
\correspondence{First and Last Authors: \email{alisia@meta.com},\email{marialomeli@meta.com}}

\begin{document}

\maketitle


\section{Introduction}
Large Language Models (LLMs)~\citep{devlin2019bertpretrainingdeepbidirectional,chowdhery2022palmscalinglanguagemodeling,brown2020language, Vaswani2017AttentionIA} have risen in popularity due to their remarkable ability to process and generate human-like text \citep{radford2018improving}.
However, the scarcity of annotated task-specific data makes it difficult to unlock new capabilities, like using tools, and to enable LLMs to solve complex tasks, e.g. tasks that require multi-step reasoning.

One solution, consisting of enriching the data with human annotations collected for specific tasks, is expensive, time-consuming~\citep{Gilardietal2023, llama2}, and prone to human errors and biases~\citep{Sylolypavanetal2023}. 

Alternatively, synthetic data that mimic real-world patterns can be constructed for fine-tuning, but ensuring the factuality and fidelity of the generated data remains challenging~\citep{liu2024bestpracticeslessonslearned}. Furthermore, constructing synthetic data for complex tasks that involve multi-step reasoning or that require aggregating complex information, for example, leveraging a table with SQL as the intermediate step for answering a question, is an open research question. If a general recipe for creating instruction-tuning synthetic data is used, only a small fraction of synthetic examples are suitable for multi-hop question-answering, and these can exhibit poor quality~\citep{chen2024essentialfactorscraftingeffective}.

We propose ~\ourtool{}, a self-augmentation and curation approach that produces \textit{high quality synthetic data grounded in real-world sources} for two complex tasks. By basing the data generation process on real-world sources,~\ourtool{} steers the examples to be more realistic, diverse, and factually correct. The curation step filters out low-quality data. The importance of data quality in enabling instruction-following capabilities has been demonstrated~\citep{zhou2023limaalignment}, it is equally important to enable other capabilities in LLMs. 

In this paper, we propose~\ourtool{}, a method that generates synthetic data for fine-tuning LLMs, unlocking two advanced capabilities: 1) \textit{question answering with tables}, which requires aggregating information from a table using SQL to answer the question and 2) \textit{multi-hop question answering}, where each question requires finding and reasoning over multiple supporting documents to answer
it. Each of these tasks depends on a distinct type of data: the first utilises tables, while the second relies on a collection of documents. Models fine-tuned with~\ourtool achieve improved performance without relying on human annotations, providing evidence that this is an effective data generation method.

\ourtool{} consists of three stages: \textit{Dataset Generation}, \textit{Dataset Curation}, and \textit{Model Fine-tuning}, see~\autoref{method_pipeline}. At the \textit{Dataset Generation} stage, we select a data source (such as tables found on the web or articles from Wikipedia) to \textit{ground} the generation in realistic information. Then, our method picks an \textit{attribute or seed} to trigger the generation and condition its components, for instance, an entity in a Wikipedia article or a factual statement about a table.
Given the attribute, the method then produces the full example: the question, the reasoning chain (e.g., the steps for multi-hop question answering, or an SQL statement) and an answer.

At the \textit{Data Curation} stage, the augmented dataset is split in two sections: the first half is used to fine-tune a LLM. This intermediate fine-tuned model is used to curate the second section of the synthetic dataset through an imputation step followed by a filtering step using rejection sampling.

For the latter, we reject examples for which the intermediate fine-tuned model fails to produce the correct answer in $k$ tries. This provides a high-quality curated dataset for fine-tuning, which results in a better performing model on a given task.

To summarise, our key contributions are: 1) a novel, scalable approach for producing synthetic data based on real data sources, showcased on two challenging tasks, and 2) a curation method that leverages slice training, filtering, and imputation to produce higher quality data and enhanced task performance.

\section{Related Work}

\textbf{Synthetic Data Generation using LLMs} 
A number of works leverage language models to generate synthetic datasets suitable for pre-training, fine-tuning or instruction-tuning. Some rely on knowledge-probing, by generating a continuation or predicting missing words in a close-style template~\citep{Schick2020FewShotTG,schick2021exploitingclozequestionsshot,Petroni2019LanguageMA,Jiang2019HowCW}. 
Other works improve the quality of synthetic data by using different model-based or human filtering techniques~\citep{Schick2021GeneratingDW, liu-etal-2022-wanli,li2024selfalignmentinstructionbacktranslation,thoppilan2022lamdalanguagemodelsdialog,Schimanski2024}. Another line of work consists of generating high-quality synthetic data for general instruction tuning \citep{Yinetal2023dynosaur,zhou2023limaalignment,wang2023selfinstructaligninglanguagemodels,alpaca,honovich-etal-2023-unnatural,wang-etal-2022-super,wang-etal-2023-self-instruct} where the LLM generates diverse potential tasks
based on the valid fields in the given dataset. However, it is hard to synthetize high-quality multi-hop instruction-tuning data. If a general recipe is used~\citep{wang-etal-2023-self-instruct}, only a small fraction of synthetic instruction-tuning samples are multi-hop, and they can exhibit poor quality~\citep{chen2024essentialfactorscraftingeffective}.
For obtaining data for tabular question-answering task, previous approaches typically rely on human annotations~\citep{li2023llm}. 

\ourtool{} creates synthetic datasets for two complex tasks: tabular question answering and multi-hop question answering. It leverages real data to identify a task-specific seed, which steers the examples to be more realistic, diverse, and factually correct.
Since it uses the LLM itself, it boosts the quality of the data, avoiding human-in-the-loop steps.

Some recent works also leverage real-world documents from the web~\citep{nguyen2024betteralignmentinstructionbackandforth,Ziegler2024CRAFTYD}, open-source code snippets to generate diverse instruction data for code generation~\citep{wei2024magicoderempoweringcodegeneration,dubey2024llama3herdmodels} or metadata from existing datasets to create instruction-tuning data~\citep{Yinetal2023dynosaur}. In contrast, ~\ourtool{} does not require a back-translation approach or initial fine-tuning to generate the task-specific seed.

\textbf{Teaching LLMs to Use Tools}
Enabling LLMs with tool-use extends their abilities to manipulating structured data, retrieving information from external sources, or interacting with APIs\citep{parisi2022talmtoolaugmentedlanguage, schick2023toolformer, Tang2023ToolAlpacaGT}. 

In most approaches, tool usage is restricted to inputs that are strings or numbers. However, a large amount of information is currently stored in relational databases. Unlocking the ability for LLMs to compose SQL queries has unlimited potential, but producing SQL queries that are relevant for a given database schema remains challenging. See Appendix~\ref{extended_related_work} for an extended related work section.

\section{Method}
\begin{figure*}[ht]
\centering
\scalebox{0.9}{
\includegraphics[width=\textwidth]{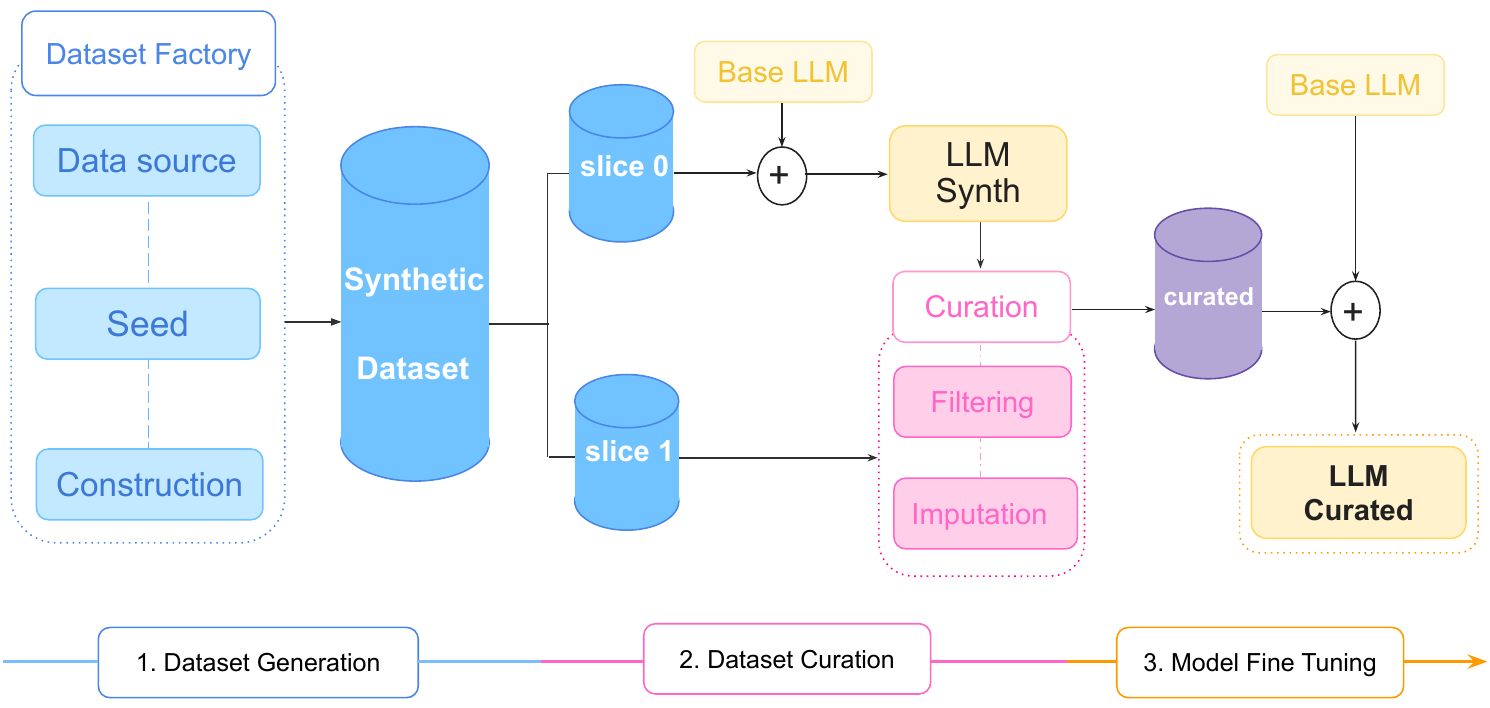}}%
\vspace{-4mm}
  \caption{\textbf{\ourtool{} Method.}  At \textit{Dataset Generation}, we choose a data source to build the dataset. For each example, we select a seed or attribute to condition the generation, and use the data source entry and seed for each example. The resulting synthetic dataset is sliced in two: slice 0, used to fine-tune an intermediate version of the LLM(\FTone{}), which is then used to curate slice 1 through filtering and/or imputation during \textit{Dataset Curation} step. At \textit{Model Fine-tuning} stage, the final LLM (~\FTtwo{}) is trained on the curated synthetic dataset.}
  \label{method_pipeline}
\end{figure*}

\ourtool{} provides a way to generate high-quality synthetic data leveraging self-augmentation and self-improvement using real-world data as sources. The generated synthetic examples can be used for fine-tuning a LLM. 

\ourtool{} consists of three stages: \textit{Dataset Generation}, \textit{Dataset Curation}, and \textit{Model Fine-tuning}. In particular, we present how to generate synthetic data suitable for two challenging tasks: multi-hop question answering (MHQA), using documents, to test reasoning; and tabular question answering (TQA), using tables, to test tool use.

\textbf{Multi-hop question answering}
In MHQA, we generate a synthetic dataset consisting of multi-hop question-answer pairs, augmented with the reasoning chain used to generate the answer. See Figure~\ref{mhqa_example} for an overview of the procedure. 

\textbf{Tabular question answering}
In TQA, we generate a question-answer dataset where each question is based on a table as the data source. In this case, the synthetic examples are
augmented with an SQL statement built from automatically-generated facts based on the initial table. See Figure~\ref{tqa_example} for an overview of the procedure for TQA.

\subsection{Dataset Generation}
\label{sec:data_gen}
\subsubsection{Data source selection} We first select a data source, which can be an existing dataset re-purposed for the task, a collection of existing data points, or structured information (graphs, tables). There is no need for human annotations on the entries, as~\ourtool{} enriches the data by self-augmentation.

\textbf{MHQA} We use English Wikipedia \citep{wiki} as the data source, since it contains articles in natural language and additional meta-information such as links to related articles. 
Firstly, we randomly select an initial article, denoted as $D_1$, among all available Wikipedia articles. For each article $D_1$, we collect a pool of $n\geq2$ related articles, from which we sample a second document $D_2$.

\textbf{TQA} We use four thousand unlabeled tables from the WikiSQL~\citep{zhongSeq2SQL2017} training dataset as sources.

\subsubsection{Seed} To create a synthetic example, we first generate a task-specific \textit{seed or attribute} chosen at random from each entry in the source data. The seed anchors the creation of the entry, making it consistent throughout the generation process. 

\textbf{MHQA} A \textit{MHQA seed} corresponds to an entity $E$ sampled from the article $D_1$. $E$ is also the ``hop'' in the multi-hop question $Q$, since $E$ links the $n=2$ sub-questions in which $Q$ is decomposed. For example, in
Figure~\ref{mhqa_example}, we randomly sample the article $D_1=$"The Moon" and the entity $E=$"Apollo 11", $E$ is the MHQA seed. Then, we pick $D_2=$"Neil Armstrong" from the pool of related articles, since it contains a paragraph where "Apollo 11" is included.

\begin{figure}[h!]
\centering
\scalebox{1}{%
\includegraphics[width=\columnwidth]
{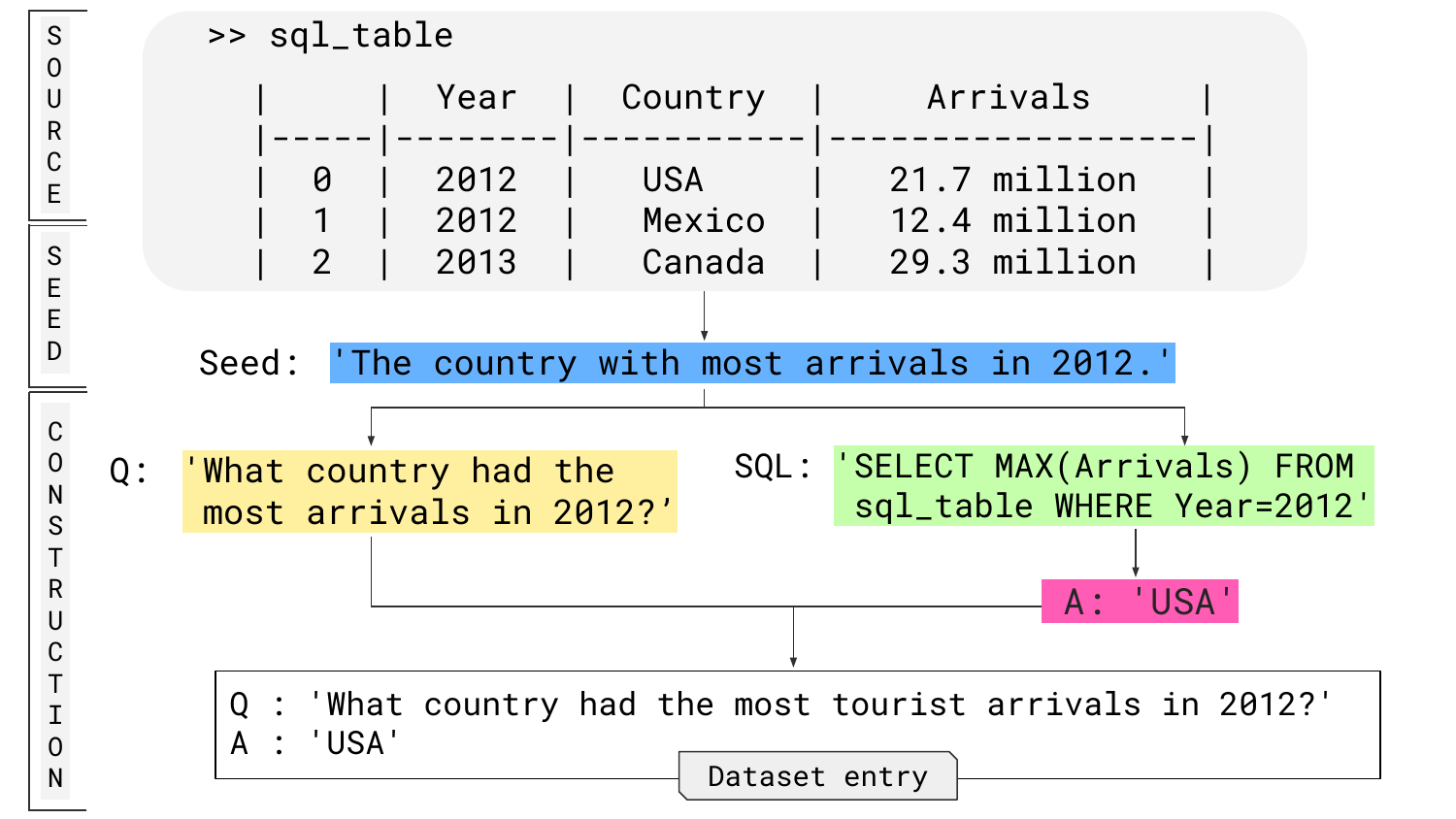}}
 \vspace{-2mm}
\caption{\textbf{
  TQA synthetic data generation process.} We generate the seed: a fact based on the table. Given the seed and table, an \textit{SQL query} is generated as well as its translation into natural language \textit{Q}. Then, the SQL is executed to obtain the answer \textbf{A}.}
  \label{tqa_example}
\end{figure}

 \textbf{TQA} A \textit{TQA seed} corresponds to an interesting fact derived from the table when prompting an instruction-tuned language model. For example, in~\autoref{tqa_example}, the TQA seed is "The country with most arrivals in 2012". See~\autoref{fig:tqa_inspiration} for the prompt. 

\subsubsection{Dataset construction} Chain-Of-Thought~\citep{Weietal2022COT} prompting has been used to break a complex task in simpler steps to make it easier to solve by a LLM. 

Analogously, we use the task-specific seed to build synthetic step-by-step data, decomposing the generation into intermediate steps. 

\begin{figure*}[ht]
\centering
\scalebox{0.9}{
  \includegraphics[width=\textwidth]{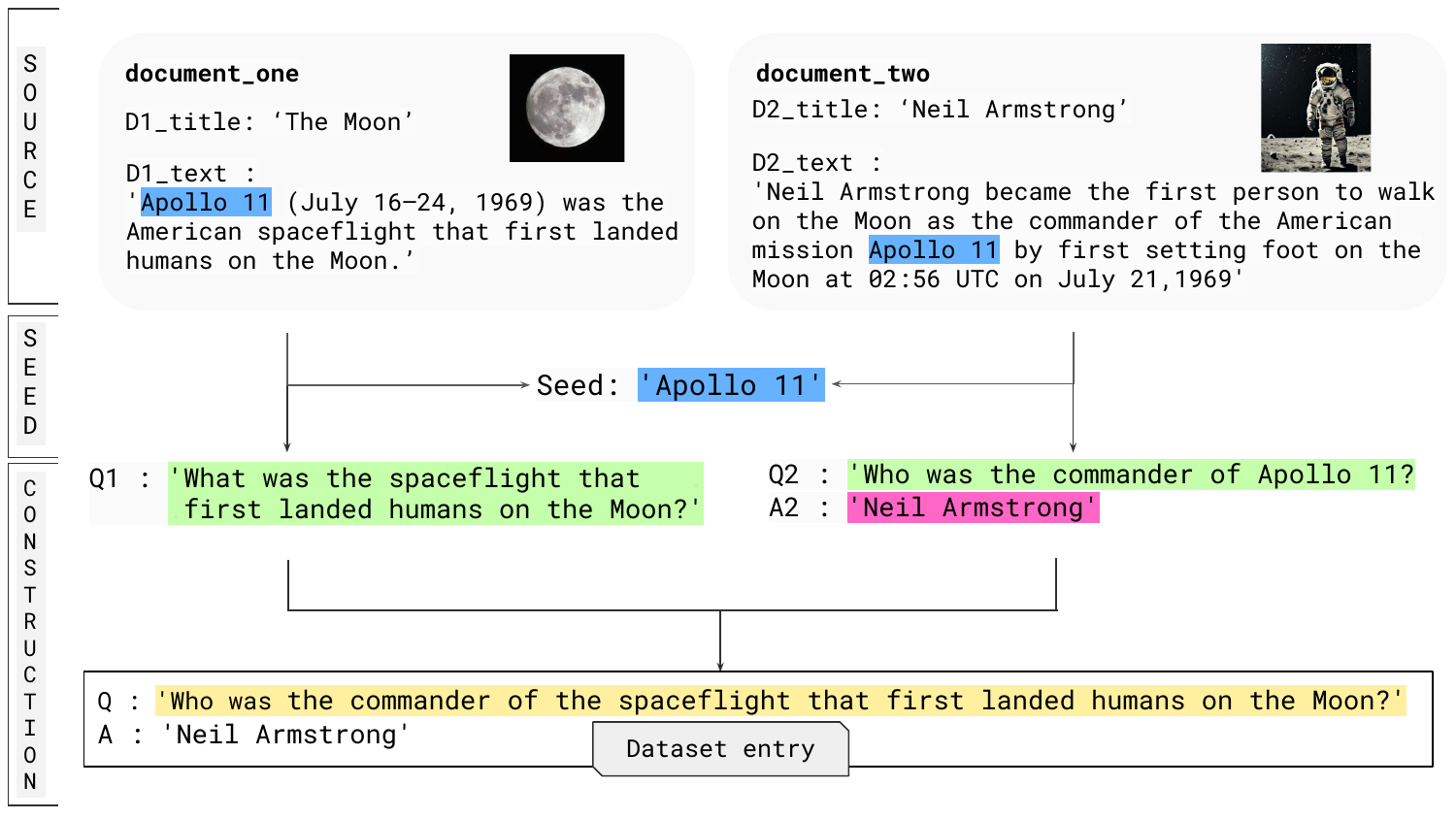}
  }
  \vspace{-4mm}
  \caption{\textbf{MHQA synthetic data generation process:}. We randomly pick an article $D_1$, e.g. "The Moon". We select the \textit{MHQA Seed} by retrieving an entity E from $D_1$'s entities e.g. "Apollo 11". We then sample from $D_1$'s pool of related articles such that $E$ is present e.g. we select
  $D_2$ titled "Neil Armstrong". A question $Q_1$ is generated from $D_1$, such that its answer $A_1$ is the entity $E$. A second question $Q_2$ is generated from $D_2$, such that its main topic is the entity $E$. We then prompt an LLM to merge the two questions into a multi-hop one $Q$, based on the common entity. The training example comprises of $Q$, $A$, the sub-questions, reasoning chain, and the entity.} 
  \label{mhqa_example}
\end{figure*}

\textbf{MHQA} We prompt an instruction-tuned language model to generate two questions: a question $Q_1$ based on $D_1$, whose answer is the selected entity $E$; and a second question $Q_2$, based on $D_2$ such that its main topic is $E$. 
See Figures~\ref{q1_gen_app} and~\ref{q2_gen_app} for the prompts. For example, in Figure~\ref{mhqa_example}, $Q_1=$ "What was the spaceflight that first landed humans on the Moon?", the hop is $E=$ "Apollo 11" and $Q_2 =$ "Who was the commander of Apollo 11?". 
We prompt the LLM to merge the two questions to generate the two-hop question $Q$ using the entity as a conceptual link (hop). See~\autoref{fig:merge_mhqa} for the prompt. 

\textbf{TQA} We zero-shot prompt the LLM to generate an SQL statement providing the table and the TQA seed as context, see~\autoref{fig:tqa_sql} for the prompt. The generated SQL statement is executed using the \textit{sqlite3}\footnote{https://www.sqlite.org} Python library to obtain the answer formatted as a table. If the generated statement is invalid, we discard it.

\subsection{Dataset Curation}
During curation, the dataset is divided into two sections, each of which contains half the number of synthetic examples. The first section is used to fine-tune an intermediate LLM~( denoted by~\FTone{}). \FTone{} is used to self-improve the quality of the second section of the data using an imputation and filtering steps (in purple in~\autoref{method_pipeline}).

\subsubsection{Data filtering} The filtering step consists of using the fine-tuned model \FTone{} to predict the output of the question in the synthetic example for $k$ tries. 
If the output cannot be predicted, it is assumed that the example is low quality and is not included in the final curated dataset.

\textbf{MHQA and TQA} In both cases, we check if the predicted answer given by \FTone{} matches the original answer in the synthetically generated example. If after $k = 3$ tries the model has not provided the correct answer, we discard the entry. 

\subsubsection{Data imputation} The imputation process involves discarding parts of the augmented data points and using~\FTone{} to fill in the blanks.

\textbf{MHQA} We discard $Q_1$ and provide~\FTone{} with $Q$, $Q_2$, $E$, and the sampled document $D_1$ as context and ask it to reconstruct $Q_1$. The new candidate $Q_1'$ for $Q_1$ is then assessed: if $A'$ (the answer to the new multi-hop question $Q'$ resulting from assembling $Q_1'$ and $Q_2$) matches $A$ (the original answer to $Q$) then we keep the example. We find that asking the model to reconstruct parts of the multi-hop question in-context results in a more natural and cohesive question, removing some of the unnaturalness of the text. In Appendix~\ref{sec:imputation}, we quantify how natural the generated MHQA questions are by measuring perplexity before and after imputation.

\textbf{TQA } the curation process consists only of the filtering step. 

In MHQA, the curation step removes around 13\% of the originally generated questions. In TQA, we keep 27\% of the original examples.

\subsection{Model fine-tuning}
\begin{figure*}[ht]
\centering
\scalebox{0.8}{%
  \includegraphics[width = \textwidth]
  {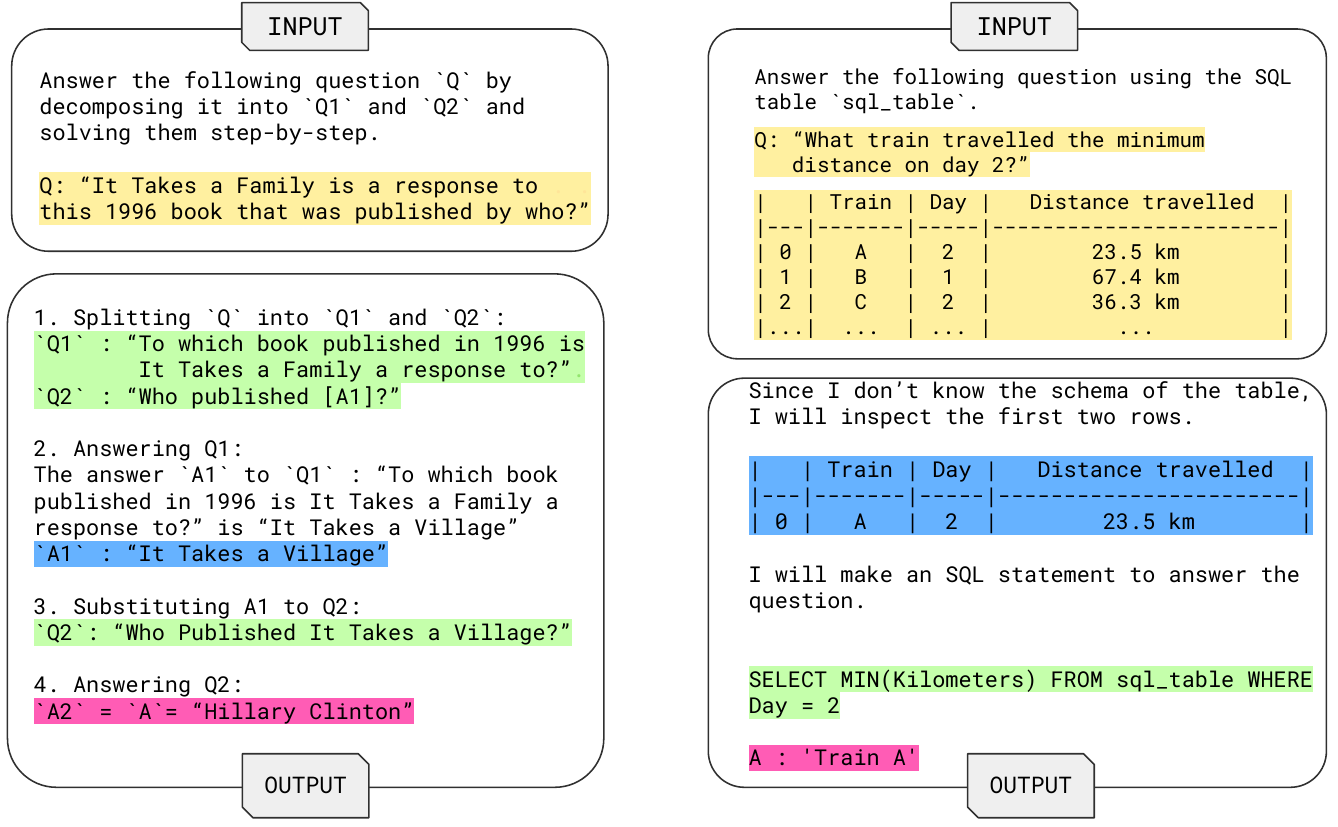}%
  }
\vspace{-4mm}
  \caption{\textbf{MHQA (left).} Model's response to a multi-hop input question (in yellow). The colours correspond to the generation of the augmented entries: the decomposition into sub questions, (in green), the seed (in blue), and the final answer (in pink).\\
  \textbf{TQA  (right).} Model's response to a tabular input question (in yellow). The coloured parts correspond to the generation of the augmented entries: SQL (in green), the seed, (in blue), and the final answer (in pink).
  \label{fig:application_inference}
  }
\end{figure*}

We can fine-tune a pretrained or an instruction-tuned LLM with the~\ourtool{} synthetic dataset. We use our dataset for supervised training of both the reasoning chain and the final answer. The resulting~\FTtwo{} model is equipped with the relevant capability for the task of interest. See~\autoref{fig:application_inference}, for an example response from the model fine-tuned with the~\ourtool{} approach for MHQA and TQA.

\section{Experimental Setup}

\ourtool{} has potential in many important domain-specific applications, such as medical or legal QA, where accessing large quantities of annotated data is costly or impossible due to proprietary constraints. In this paper, for reproducibility purposes, we focus on publicly available data and evaluation benchmarks. Each of these benchmarks contains source data for one of the two tasks of interest: \textit{tabular question answering}, and \textit{multi-hop question answering}. For MHQA, we use the HotpotQA~\citep{Yang2018HotpotQAAD} benchmark and for TQA, we use the WikiSQL~\citep{zhongSeq2SQL2017} benchmark. We compare our~\ourtool{} method with a number of baselines. 

\subsection{Multi-Hop QA Experimental Setup}

\textbf{Evaluation data}  The HotpotQA benchmark~\citep{Yang2018HotpotQAAD} (HPQA) consists of 113,000 examples of multi-hop question-answer pairs based on Wikipedia, split in train, test, and validation sets.
 \citet{Yang2018HotpotQAAD} recommend the FullWiki setup to test multi-hop reasoning abilities of models, for this reason, we use this setup for the evaluation with the test set.
 
Each entry in HPQA is constructed such that:
1) each question requires finding and reasoning over multiple supporting documents to be able to answer it; 2) each entry provides sentence-level supporting facts for strong supervision and explainability of the prediction; 3) each question is either comparison or bridge type. 

A comparison question entails comparing the same concept between $n$ objects, e.g. "Who is the tallest student in class?", while a bridge question builds on a logical and/or causal link and requires deriving statements to get to the answer. E.g. "What is the height of the student that topped the entry exam?" , which requires first identifying the student that topped the exam. The hop length is the number of compared objects, for comparison questions, and for bridge questions, it is the number of links. We choose $n=2$ to be consistent with HPQA. The test set consists of $7,405$ entries, split evenly between bridge and comparison questions.~\ourtool{} only generates synthetic data for bridge questions. To counterbalance this, we include five hundred comparison questions from HPQA training dataset in the fine-tuning dataset.

\textbf{Evaluation data contamination checks} For each synthetic question generated by~\ourtool{}, we check if its entity $E$ (seed) is present in any of the questions in HPQA's test-set. If so, we check whether the two questions are the same. We found that none of the synthetic data overlaps with the questions in HPQA test set.

\textbf{Metrics} We measure the performance using the soft exact match metric (soft-EM). Soft-EM uses string comparison, it is one if the generated output contains the golden answer and zero otherwise. 

\textbf{Model} In the MHQA experiments, we use the Llama2 70B-Chat LLM, we fine-tune it with~\ourtool{} and compare it with other baseline methods.~\FTtwocomb{} denotes the fine-tuned model that uses~\ourtool{} with 1250 synthetic examples in addition to 500 examples from the HPQA training set.~\FTtwo{} denotes the fine-tuned model that uses~\ourtool{} with 1250 synthetic examples only. The 1250 synthetic examples consist of bridge questions only and are generated using a collection of 50 randomly selected Wikipedia articles.

\textbf{Baselines} We compare~\ourtool{} against the following baselines (for all listed models, we use two different prompt templates: a zero-shot and a three-shot CoT, see~\autoref{ft2_mhqa} in Appendix~\ref{sec:appendix_prompts_baselines} for details): \\
\textit{Instruction-tuned LLMs}: we include both Llama2 70B-Chat and Claude3.5 Sonnet~\citep{claude}.\\
\textit{Fine-tuned LLM (HPQA  only)}: we fine-tune Llama2 70B-Chat model with 500 examples from the HPQA training split.\\
\textit{\FTone{} (Synthetic dataset only)}: we fine-tune Llama2 70B-Chat model with 1250 synthetic examples from Slice 0 (see~\autoref{method_pipeline}),  {\em without} the data curation step.\\
\textit{\FTonecomb{} (Synthetic and HPQA)}:   we fine-tune Llama2 70B-Chat with the uncurated synthetic data in addition to the 500 HPQA examples. 

\subsection{Tabular QA Experimental Setup}

\textbf{Evaluation data} The WikiSQL benchmark~\citep{zhongSeq2SQL2017} consists of 80,654 hand-annotated examples of natural language questions, SQL queries, and tables created from 24,241 tables extracted from Wikipedia. The validation split contains 7,857 examples after removing non-executable SQL tables, see Appendix~\ref{sec:SQL-non-executable} for more details. 

\textbf{Evaluation data contamination checks} During the~\ourtool{} synthetic dataset construction, we select the tables in the train split of the WikiSQL benchmark as the source. Since we evaluate with the test set, there is no overlap or evaluation data contamination since these sets are mutually exclusive by design.

\textbf{Metrics} We measure performance using the exact match (EM) and the soft-EM  metrics. The EM metric uses string comparison, it equals one if the golden answer is equal to the generated answer and zero otherwise.

\textbf{Model} For TQA, we use the Starchat-beta language model~\citep{li2023starcodersourceyou} as the initial language model (and use the following parameters for finetuning: batch size 32, 100 steps, learning rate of 0.0001 and linear warm-up). The Starchat model is an instruction-tuned LLM with 16 billion parameters trained to act as a helpful coding assistant. This model is a fine-tuned version of StarCoder~\citep{li2023starcodersourceyou}, that consists of a LLM pre-trained on a large code corpus, containing SQL statements, and fine-tuned on 35B Python tokens. We generate ten thousand SQL statements based on the source tables and keep eight thousand examples per slice.

\textbf{Baselines} We compare the performance of~\ourtool{} method against a variety of baselines consisting of prompting the Starchat-beta instruction-tuned language model: \\
\textit{Zero-shot Table QA~(\autoref{fig:zero_shot_table_qa})}: zero-shot prompt with task instruction, table and question. \\
\textit{One-Shot No Context QA~(\autoref{fig:one_shot_no_context_qa})}: one-shot example with a question and answer prompt with task instruction and the actual question to answer. \\
\textit{One-Shot Table QA~(\autoref{fig:one_shot_table_qa})}: prompt including the table for the one-shot example and the question to answer.\\
\textit{One-shot Table+SQL QA~( \autoref{fig:one_shot_table_sql_qa})}: the prompt includes an example containing the table and question, and an instruction suggesting that the model can leverage an SQL tool. We then execute the predicted SQL to obtain the answer.\\
\textit{\FTone}: Fine-tune the model with synthetic data {\em without} applying the data curation step.

\section{Results}
\label{results}
\subsection{Multi-Hop question answering}

\begin{table*}[ht]
  \centering
  \caption{\textbf{Evaluation of \ourtool{} on Multi-hop question answering.}~\FTtwocomb{} and~\FTonecomb{} are fine-tuned with 500 entries from HPQA and 1250 entries from the~\ourtool{} synthetic data.~\FTone{} and~\FTtwo{} are fine-tuned with 1250 entries from the~\ourtool{} synthetic data only.}
  \label{tab:toolcurator_results-mhqa}
   \scalebox{0.9}{
  \begin{tabular}{l|r|r}
    \toprule
    \textbf{Method} & \textbf{0-shot} & \textbf{3-shot CoT prompt}\\
    \midrule
    Llama2 70B-Chat         & 40.45\% & 44.13\% \\ 
      Claude 3.5 Sonnet  & 50.3\% & 53.4\% \\ 
Fine-tuned LLM (HPQA data only)                   & 53.22\% & 58.40\% \\
LLMSynth (synthetic data only (bridge questions)) & 52.31\% & 56.70\% \\ 
LLMSynth-datamix (HPQA and synthetic data (bridge questions)) & 57.46\% & 62.73\% \\ 
LLMCurated (synthetic data only (bridge questions))   & 64.07\% & 64.68\%\\
\textbf{\FTtwocomb{}  (HPQA and synthetic data (bridge questions))} 
 & \textbf{65.23\%} & \textbf{66.05\%} \\
    \bottomrule
  \end{tabular}}
\end{table*}

We report the experimental results in Table~\ref{tab:toolcurator_results-mhqa} using zero-shot and three-shot prompts (~\autoref{ft2_mhqa}). All fine-tuned models outperform the first two baselines where we prompt the Llama2 70B-Chat and Claude3.5 Sonnet instruction-tuned models. We observe that using only synthetic data or only HPQA data for fine-tuning has worse performance than when combined, whether or not the synthetic data is curated (\FTtwo{},~\FTone{}). In all models where we use the full~\ourtool{} pipeline we see further performance improvements (~\FTtwo,~\FTtwocomb) vs not curating the data (~\FTone,~\FTonecomb).

In~\autoref{tab:toolcurator_results-mhqa}, we can simulate the no-data regime by analysing the case where we fine-tune Llama2 70B-Chat with synthetic data only. We observe that fine-tuning without the additional 500 entries from HPQA minimally hinders performance (since the difference between~\FTtwo vs~\FTtwocomb is 1.13\%).

\begin{figure}[h!]
\centering
\scalebox{0.9}{%
\resizebox{\columnwidth}{!}{
  \includegraphics[width=10cm]{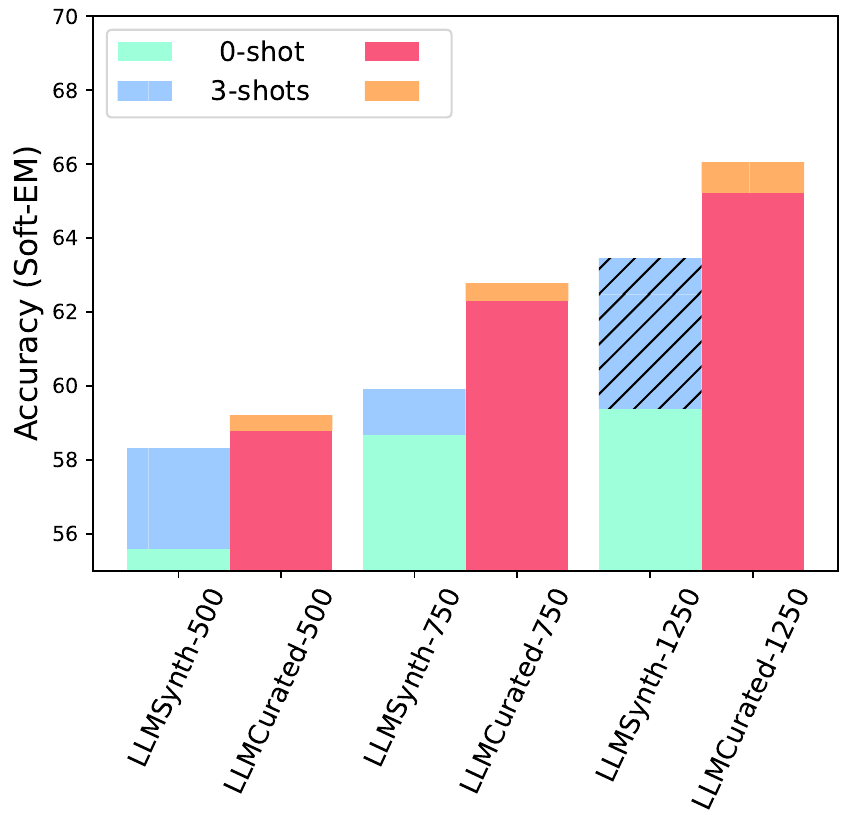}}
  \vspace{-12mm}
  }\caption{\textbf{Synthetic Data scaling performance.}
  \ourtool{} performance when increasing the amount of synthetic data in the data mix, before and after curation. All models in the plot use the 500 HPQA examples but vary the number of synthetic examples, for example, LLMSynth-500 uses 500 HPQA and 500 synthetic examples.
  \label{scaling_law_plot}
  }
\end{figure}

\begin{table*}
\centering
\caption{\textbf{Tabular question answering.} The models~\FTone{} and~\FTtwo{} are fine-tuned using~\ourtool{} curated synthetic data only. Performance comparison on the WikiSQL test set.}
\label{tab:toolcurator_results_SQL}
 \scalebox{0.9}{
\begin{tabular}{l|r|r}
\toprule
\textbf{Method} & \textbf{Exact Match} & \textbf{~~~~~Soft-EM} \\ 
\midrule
Starchat-beta (one-shot no context QA)& 0.25\% & 16.22\% \\
Starchat-beta (zero-shot table QA)& 1.83\% & 20.07\% \\ 
Starchat-beta (one-shot table QA)& 2.03\% & 31.06\% \\ 
Starchat-beta (one-shot table+SQL tool QA)& 12.30\% & 34.13\% \\ 
LLMSynth (synthetic data only)& 23.86\% & 34.21\% \\
\textbf{\FTtwo{} (synthetic data only)}& \textbf{34.50\%} & \textbf{42.80\%} \\ 
\bottomrule
\end{tabular}}
\end{table*}

\textbf{Scaling performance}~\ourtool{} can be leveraged when the amount of available data is low. In Figure~\ref{scaling_law_plot}, we study how performance changes when adding more synthetic data in the fine-tuning data mix, which includes 500 samples from the HPQA train split. We perform the analysis on both~\FTonecomb{} and~\FTtwocomb{} to showcase the impact of the curation technique.
During the curation step, the following percentages of samples were removed: 7\% for 500, 8\% for 750, 11\% for 1250.

In both cases, uniformly over all data mix sizes, we see that applying the~\ourtool{} pipeline results in a stronger model.
For the~\FTonecomb{} model (fine-tuned with uncurated samples), providing more synthetic examples leads to a steady improvement in performance across all data sizes, for both zero-shot and three-shot prompting variants.
The~\FTtwocomb{} model follows a similar trend, consistently outperforming the uncurated version of the model.
Overall, we observe that using our synthetic data generation pipeline to construct more synthetic data brings further performance gains.

\textbf{Analysis of performance with respect to different question types and levels of difficulty}
In Appendix~\ref{sec:further_exps_bridge_vs_comparison}, we study the capabilities of our model by analysing the performance of LLM-Curated-1250 with a particular focus on the type and difficulty of the questions: hard/medium/easy bridge and comparison questions. In~\autoref{tab:bridge_and_comparison_split}, 
we compare the performance of the base model, the model fine-tuned with HPQA only, and the model fine-tuned using~\ourtool{}, according to the difficulty level, provided in the HPQA train dataset. We also subdivide the results according to question-type (bridge vs. comparison). 

We observe that~\ourtool{} performs better in all types of questions and difficulties, with an average absolute performance gain of +13.38\% on the base LLM and an absolute performance gain of +7.35\% compared to the LLM fine-tuned on HPQA. In particular, by applying our method, the resulting model achieves absolute performance gains of +16.8\% and +16.5\% on hard bridge and comparison questions, respectively, compared to the baseline. Furthermore, it is interesting to see substantial improvement in answering comparison-type questions despite not generating these during dataset construction.

\textbf{Extending~\ourtool{} data-generation method to comparison-type questions} For comparison questions, we require to compare the same concept between two objects. The dataset generation method from Section~\ref{sec:data_gen} can be extended to comparison-type questions as follows: For the first document, we select the entities that have named entity recognition (NER) properties such as date, location, nationality, among others. We then select a second document and from its pool of entities, we draw an entity that has the same NER property as the first entity from the first document. We can then prompt a LLM to produce a question that compares the common NER property between both documents.

In~\autoref{tab:toolcurator_results-mhqa_cont}, we show that if we use the~\ourtool{} approach with synthetic data consisting of both bridge and comparison questions, then,~\FTtwo{} has comparable performance to~\FTtwocomb{} from~\autoref{tab:toolcurator_results-mhqa} which uses comparison questions from the HPQA dataset and synthetically generated bridge questions (1.13\% difference).

\begin{table*}[ht]
  \centering
  \caption{\textbf{Evaluation of \ourtool{} on Multi-hop question answering.} The models shown are fine-tuned with 1250 entries from the \ourtool{} synthetic data for both comparison and bridge questions.}
  \label{tab:toolcurator_results-mhqa_cont}
   \scalebox{0.9}{
  \begin{tabular}{l|r}
    \toprule
    \textbf{Method} & \textbf{0-shot} \\
    \midrule
  LLMSynth (synthetic data (bridge and comparison questions) & 56.01\% \\ 
\textbf{\FTtwo{} (synthetic data only (bridge and comparison questions)}   & \textbf{64.5\%} \\
    \bottomrule
  \end{tabular}}
\end{table*}

\textbf{Effectiveness fine-tuning smaller LLMs (Llama3 8B-instruct and Llama4 17Bx16E)} In Appendix~\ref{sec:non_monolithic_setting}, we fine-tune Llama3 8B-instruct and Llama4 17Bx16E on 1250 synthetically-generated examples with~\ourtool{} using Llama2 70B-Chat for the dataset generation. We showcase that our method remains effective when fine-tuning smaller LLMs, resulting in an accuracy increase of 23.06\% for the Llama3 8B-instruct LLM and 36.90\% for the LLama4 17Bx16E models respectively. 

\textbf{Effect of not using data grounded in real-world sources} In Appendix~\ref{sec:further_exps_ungrounded_data_only}, we apply~\ourtool{} starting from an ungrounded synthetic dataset to show the benefit of real-world data as a source. Namely, if we use ungrounded synthetic data to fine-tune either  Llama3 8B-instruct or Llama2 70B-Chat, the resulting models have an accuracy loss of -7\% on average.

\subsection{Tabular question answering}

The experimental results for TQA are shown in~\autoref{tab:toolcurator_results_SQL}.
Providing no context about the table when prompting the instruction-tuned StarChat language model has very poor performance (first row). This is expected, since the questions in WikiSQL require information contained in the table, while the model does not have any other information except for the general knowledge stored in its parameters. Even if we pass the table in the context, the performance does not improve much due to the inherent difficulty to ingest structured data. We observe that passing the table in a zero-shot fashion is not good (second row). Even including an example of table usage in a one-shot fashion (third row) improves only the soft-EM but the EM metric remains low (2.03\%). 
Performance increases once we enable the model to use the SQL tool and provide a one-shot example containing both the relevant table and the SQL query (fourth row, $EM=12.3\%$). 

We observe an increase in performance if we use~\ourtool{} to fine-tune the StarChat model (\FTtwo{}, last row). Indeed,~\ourtool{} performs significantly better than fine-tuning the StarChat language model with uncurated data only (\FTone{}, second to last row). Even so,~\FTone{} still outperforms the other baselines.

\section{Conclusion}

We introduce~\ourtool{}, a new method for generating high-quality synthetic data for fine-tuning LLMs to unlock two advanced capabilities: 1)\textit{question answering with tables} and 2) \textit{multi-hop question answering}. Our method guides and grounds the generation, taking into account real data sources and filtering for low-quality samples. We believe our method is valuable in domains where unstructured data is available as a source—such as the legal and medical fields—even though this data is typically not readily available in the form of question-answer pairs. Leveraging the~\ourtool{} method bypasses the need to use human annotators to transform such unstructured data into QA pairs. We show that fine-tuning on high quality examples produced with~\ourtool{} consistently leads to substantial improvements while using less data. We see our method as a first step towards building high-quality automatic data generation methods without human input.

\bibliographystyle{assets/plainnat}

\bibliography{arxiv_refs}

\newpage
\onecolumn
\appendix

\section{Limitations}
\label{sec:limitations}
\begin{itemize}
\item \textbf{Two hops for MHQA} The MHQA number of hops is restricted to two in this paper. However,~\ourtool{} can be extended to a number of hops greater than two. This can be done
by looping the dataset generation steps and feeding the result of the previous step as input to the next one. 

 \item \textbf{There exist a single table per query to obtain the answer in TQA} We use a single table per query in TQA. However,~\ourtool{} can be extended to more complex scenarios that require to first identify the relevant table from a set of tables. In order to do this, we could use table retrieval and leverage \citet{Herzig2020TAPAS} as an encoder model for tables.
\item \textbf{Multi-table tool use is not supported} \ourtool{} cannot handle queries that require to aggregate information contained in multiple tables e.g. SQL join statements.

 \item \textbf{Use of rejection sampling for MHQA dataset construction}
Our method could be improved with more clever sampling techniques beyond rejection sampling, for instance, leveraging the multihop retrieval approach of~\citet{XiongetalMHRetrieval2020}. 

 \item \textbf{\ourtool{} is restricted to question-answering tasks} We focus on two different data-types as sources, documents and tables, hence,~\ourtool{} can be extended to any domain that has such data-types as source even if it is not publicly available such as: legal and medical domains, among others. We only tackle question-answering tasks and focus on teaching the model the skills of producing reasoning chains in the MHQA format or producing SQL statements relevant to a given (source) table. We consider extending our method to other tasks and skills an interesting future direction.

  \item \textbf{Source data needs to be checked for inconsistencies} We assume that if the user uses a given data source, some initial checks are conducted to remove inconsistencies, such as checking if the tables are complete or if the webpages used are duplicated or contain misinformation. However, some inconsistencies in the data source are potentially mitigated by the filtering process, since the answerability check helps with logical soundness. In the TQA case, we discard nonexecutable SQL statements, but the tables need to be properly parsed by the sqlite library.
\end{itemize}
\section{Extended related work}
\label{extended_related_work}
\textbf{Teaching LLMs to use tools} Enabling LLMs with tool use extends their abilities to manipulating structured data, retrieving information from external sources, or interacting with APIs. Various works augment LLMs with general tools or API calls~\citep{parisi2022talmtoolaugmentedlanguage, schick2023toolformer, Tang2023ToolAlpacaGT}, possibly interleaving reasoning steps with API calls ~\citep{gao2023palprogramaidedlanguagemodels,cai2024largelanguagemodelstool, paranjape2023artautomaticmultistepreasoning}. Some works investigate the use of unseen tools at test time~\citep{paranjape2023artautomaticmultistepreasoning, mekala2024toolverifiergeneralizationnewtools}.~\citet{mialon2023augmented} and~\citet{qin2023toollearningfoundationmodels} provide an in-depth review of research on augmented language models. 

In most approaches, tool usage is restricted to inputs that are strings or numbers. However, a large
amount of information is currently stored in relational databases. Unlocking the ability
to compose SQL queries has unlimited potential, but producing SQL queries that are relevant for a given database schema remains challenging. 

\textbf{SQL for LLM tool-usage and transformers for tabular data}
A variety of benchmarks ~\citep{pasupat-liang-2015-compositional} have been proposed to evaluate the ability of the LLM to generate relevant SQL and their performance in tabular-based question answering~\citep{li2023llm, zhongSeq2SQL2017}.

Alternatively, other works adapt language models to directly handle tabular data ~\citep{herzig-etal-2020-tapas,Yinetal2020TaBERT}.~\citet{Badaroetal2023tabtransformer} provide a comprehensive overview of works that modify the transformer architecture for tabular data.

\section{Further experiments}
\subsection{Analysis of performance on different question types and levels of difficulty}\label{sec:further_exps_bridge_vs_comparison}

We study the capabilities of our model by analysing the performance of LLM-Curated-1250 with a particular focus on the type and difficulty of the questions: hard/medium/easy bridge and comparison questions. In~\autoref{tab:bridge_and_comparison_split}, we compare the performance of the base model, the model fine-tuned only on HPQA, and the model fine-tuned using~\ourtool{}, according to the difficulty level provided in the HPQA train dataset. We also subdivide the results according to the type of question (bridge vs. comparison).

\begin{table*}[t]
\centering
\caption{\textbf{Analysis of MHQA bridge and comparison questions with respect to level of difficulty.} We evaluate models on the full HPQA train dataset (where questions are labelled with easy, medium and hard). \ourtool{} outperforms the baseline and the fine-tuned on HotpotQA model, yielding a LLM  capable of handling hard questions of both types using 1250 synthetic examples.} 
 \label{tab:bridge_and_comparison_split}
  \scalebox{0.9}{
\begin{tabular}{lrrr@{\hphantom{10}}|@{\hphantom{10}}rrr}
\toprule
& \multicolumn{3}{c}{\textit{Bridge}} & \multicolumn{3}{c}{\textit{Comparison}} \\
\textbf{Model} & \textbf{Hard} & \textbf{Medium} & \textbf{Easy} & \textbf{Hard} & \textbf{Medium} & \textbf{Easy} \\ 
\midrule
Llama2 70B-Chat  & 14.5\% & 27.2\% & 30.1\%	& 66.6\% &		71.3\%	& 73.2\%\\
Fine-tuned LLM (HPQA data only)  & 20.1\%	&		29.8\%	&		34.3\%	& 74.5\%	&		78.3\%	&		82.1\% \\
LLMCurated (Synthetic data only) & 27.6\%	&		32.3\%	&		36.2\%	& 79.1\%	&		82.3\%	&		88\% \\
\textbf{\FTtwocomb{} (Synthetic and HPQA)}&\textbf{ 31.3\%}	&	\textbf{	35.6\%}	&	\textbf{	39.7\%	}&\textbf{ 83.1\%}	&	\textbf{	85.7\%	}&		\textbf{87.8\%} \\
\bottomrule
\end{tabular}}
\end{table*}

\subsection{Non-monolithic setting: fine-tuning smaller LLMs using a different model for data generation} \label{sec:non_monolithic_setting}
We fine-tune Llama3 8B-instruct and Llama4 17Bx16E on 1250 synthetically generated examples resulting from the~\ourtool{} pipeline and on 500 entries from HotpotQA. In both cases, we use Llama2 70B-Chat for dataset generation. For Llama3 8B-instruct we only generate synthetic bridge questions whereas for Llama4, we generate both synthetic bridge and comparison questions. For evaluation, we use the 0-shot prompt from~\autoref{ft2_mhqa} and the soft-EM as a metric. In both cases,~\ourtool{} gives a performance boost with respect to the corresponding base LLMs. For Llama3 8B-instruct, in ~\autoref{tab:Llama3_ablation}, we can see that compared to the performance of the base model,~\FTtwo{} shows an increase in accuracy of 23.06\%. For Llama4 17B 17Bx16E, in ~\autoref{tab:llama4}, we can see that compared to the performance of the base model,~\FTtwo{} shows an increase in accuracy of 36.90\%.

\begin{table*}[h!]
\caption{Performance of~\ourtool{} fine-tuning Llama3 8B-instruct using Llama2 70B-Chat for dataset generation}
  \label{tab:Llama3_ablation}
  \centering
  \begin{tabular}{l|r}
    \toprule
    \textbf{Model} & \textbf{0-shot} \\
    \midrule
    Llama3 8B-instruct        & 57.8\% \\ 
    LLMSynth (synthetic data only (bridge questions))                  & 64.46\% \\ 
    \textbf{\FTtwo{}  (synthetic data only (bridge questions))}  & \textbf{71.13\%} \\
    \bottomrule
  \end{tabular}
\end{table*}

\begin{table*}[ht]
  \centering
  \caption{Performance of~\ourtool{}fine-tuning Llama4-17Bx16E  using Llama2 70B-Chat for dataset generation}
  \label{tab:llama4}
  \begin{tabular}{l|r}
    \toprule
    \textbf{Model} & \textbf{0-shot} \\
    \midrule
Llama4-17Bx16E base LLM       & 49.6\% \\ 
    LLMSynth (synthetic data only (bridge and comparison questions))             & 58.7\% \\ 
    \textbf{\FTtwo(synthetic data only (bridge and comparison questions))} & \textbf{67.9\%} \\
    \bottomrule
  \end{tabular}
\end{table*}

\subsection{Applying \ourtool{} on made up toy data only}\label{sec:further_exps_ungrounded_data_only}
We tried~\ourtool{} starting with a fully made up toy dataset (ungrounded data means no real-world source), consisting of 1250 synthetic data points for fine-tuning plus 500 entries from HotpotQA. To ensure diversity in the generation, we ask the model to generate a question based on two topics $A$ and $B$ picked from the following list:
[“Moon”, “Ocean”, “Water”, “Wolf”, “Tides”, “Day”, “Light”, “Apple”, “United States”, “Europe”, “Roman Empire”, “Chocolate”, “Environment”, “India”, “Strawberries”, “Physics”, “Pen”, “Sugar”, “History”, “Jelly”, “Mug”, “Cat”, “Lion”, “Flower”, “Purple”, “Red”, “Stars”, “Electricity”, “Paper”, “Snow”, “Mount Everest”, “Table”, “Friendship”, “Book”, “Laptop”, “Phone”, “Mushroom”, “Hat”, “Coffee”, “Pasta”, “Island”, “Volcano”, “Storm”, “Key”, “Candle”, “Asia”, “Desert”, “Tree”, “River”]

The resulting two-hop questions are bridge-type and have lower perplexity than those generated starting from a grounded source. The model trained on ungrounded samples performs worse than the one trained on grounded ones. Comparing~\autoref{tab:ungrounded_data_llama3} to~\autoref{tab:Llama3_ablation}, we can see that the loss in accuracy in the case of Llama3 8B-instruct is -7.17\%. For Llama2 70B-Chat, comparing~\autoref{tab:ungrounded_data_llama2} with ~\autoref{tab:toolcurator_results-mhqa} the accuracy loss is -6.82\%. We also observe that there are repeating patterns in the structure of the questions generated, we hypothesize that this hinders generalization. For example, many of the synthetically ungrounded questions follow this structure: 'What / Who [Q1] and / or What [Q2]?' ” (i.e. "What is the name of the ancient mythological figure that is often depicted as being able to transform into a wolf, and is also associated with the full moon that occurs in March, which is also known as the Worm Moon?").

\begin{table*}[ht]
  \centering
  \caption{Performance of~\ourtool{} using made up (ungrounded) toy data as source, fine-tuning Llama3 8B-instruct, using Llama2-70B Chat for dataset generation.}
  \label{tab:ungrounded_data_llama3}
  \begin{tabular}{l|r}
    \toprule
    \textbf{Model} & \textbf{Accuracy} \\
    \midrule
    Llama3 8B-instruct         & 57.80\% \\ 
   LLMSynth (ungrounded toy data only)                   & 60.45\% \\ 
    \textbf{\FTtwo(ungrounded toy data only)}   & \textbf{ 66.37\%} \\
    \bottomrule
  \end{tabular}
\end{table*}
\begin{table*}[ht]
  \centering
  \caption{Performance of~\ourtool{} using made up (ungrounded) toy data as source,fine-tuning Llama2 70B-Chat, using Llama2 70B-Chat for dataset generation.}
  \label{tab:ungrounded_data_llama2}
  \begin{tabular}{l|r}
    \toprule
    \textbf{Model} & \textbf{Accuracy} \\
    \midrule
    LLama2 70B-Chat & 40.45\% \\ 
    LLMSynth (ungrounded toy data only)                   & 51.90\% \\ 
    \textbf{\FTtwo(ungrounded toy data only)}    & \textbf{59.70\%}\\
    \bottomrule
  \end{tabular}
\end{table*}

\subsection{On the impact of imputation} \label{sec:imputation}
We studied the perplexity of questions before and after imputation for 1) synthetic data generated from a grounded source like Wikipedia and 2) for synthetic ungrounded data. In both cases, the imputation step lowers perplexity and reduces the unnaturalness of the question. In~\autoref{tab:imputation_method} we report the average perplexity score and in~\autoref{fig:comparing_q_qq} we showcase an example of how imputation leads to rephrasing sentences in a more natural way.

\begin{table*}[ht]
  \centering
  \caption{Average perplexity of generated questions before and after imputation}
  \label{tab:imputation_method}
  \begin{tabular}{l|r|r}
    \toprule
    \textbf{Model} & \textbf{PPL before imputation} & \textbf{PPL after imputation} \\
    \midrule
    Synthetic grounded data         & 24.7 & 13.6 \\ 
    Synthetic ungrounded data       & 15.51        & 8.33 \\ 
    \bottomrule
  \end{tabular}
\end{table*}

\begin{figure}[h]
\begin{prompt}{Comparing Q pre- and post- imputation}

\textit{Before imputation:}\\
$Q$: "What pet did the poet and father of mathematician Ada Lovelace had when he was a student at Trinity out of resentment for rules forbidding pet dogs like his beloved Boatswain?" \\
$Q_1$: "What pet did the poet Lord Byron had when he was a student at Trinity out of resentment for rules forbidding pet dogs like his beloved Boatswain?"\\
$Q_2$: "Who is the father of mathematician Ada Lovelace?"\\ 
$E$: "Lord Byron" \\
$A$: "A bear" \\
$D_1$: "Lord Byron also kept a tame bear while he was a student at Trinity out of resentment for rules forbidding pet dogs like his beloved Boatswain."
\\
\\
\textit{After imputation:}\\
$Q'$: "What pet did the poet and father of mathematician Ada Lovelace had when he was a student at Trinity?" \\
$Q_1'$ : "What pet did the poet Lord Byron had when he was a student at Trinity?"
\end{prompt}
\vspace{-4mm}
\caption{Comparing Q pre- and post- imputation
   \label{fig:comparing_q_qq}
} 
\end{figure}

\subsection{More results on prompt engineering}

\begin{table*}[h!]
\centering
\caption[width=\columnwidth]{\textbf{MHQA using different prompts.} Llama2 70B-Chat accuracy across different prompts.The prompt for \textit{Role} is "You are a QA-robot. Answer the following question:".}
 \label{tab:prompteng}
\begin{tabular}{l|r}
\toprule
\textbf{Prompt Type} & \textbf{Model Accuracy (soft-EM, hotpotQA test set)}\\
\midrule
0-shot         & 40.45\% \\
Role         & 22.34\% \\ 
1-shot         & 26.65\% \\
Few-shots (5-shots)        & 21.83\% \\ 
Role (1-shot)        & 28.29\% \\ 
\bottomrule
\end{tabular}
\end{table*}
\section{SQL non-executable code filtering}
\label{sec:SQL-non-executable}
We discard SQL statements which cannot be executed with \textit{sqlite3}\footnote{https://www.sqlite.org}. Out of 50 tables, we generate 800 seed statements and keep 658 executable SQL statements.

\section{Prompts}
\label{sec:appendix_prompts_baselines}
\begin{figure}[h]
\begin{prompt}{Zero-shot Table QA prompt.}
Answer the following question using the table below. You may leverage an SQL tool.\\
\\
{\color{blue}{\{table\}}}\\
\\
Q: {\color{blue}{\{question\}}}
\end{prompt}
\caption{Zero-shot Table QA prompt for the TQA task.
   \label{fig:zero_shot_table_qa}
} 
\end{figure}
\begin{figure}[h]
\begin{prompt}{One-Shot No context QA prompt.} 
-- Example --\\
Q: What was the last year where this team was part of the US A-league?\\
A: 2004\\
\\
Now do the same for the following question.\\
Q: \color{blue}{\{question\}}
\end{prompt}
\caption{One-Shot No context QA prompt for the TQA task.
\label{fig:one_shot_no_context_qa}
} 
\end{figure}

\vspace{-6mm}
\begin{figure}[h]
\begin{prompt}{One-shot Table QA prompt.}
\texttt{-{}-} Example \texttt{-{}-}\\
Answer the following question using the table below.\\
Your answer should be short and concise.\\
\begin{verbatim}
Season | Team        | League_apps | Goals
1923   |Swindon Town | 55          | 3
1922   |Swindon Town | 14          | 4
1921   |Swindon Town | 24          | 11
1920   |Swindon Town | 26          | 16
1919   |Swindon Town | 20          | 10
1914   |Swindon Town | 23          | 12
1913   |Swindon Town | 24          | 18
1912   |Swindon Town | 12          | 9
1911   |Swindon Town | 20          | 16
1910   |Swindon Town | 30          | 19
1909   |Swindon Town | 33          | 19
1908   |Swindon Town | 34          | 28
1907   |Swindon Town | 30          | 17
\end{verbatim}
~~\\
Q: How many league appearances were there between 1907 and 1909 (inclusive)?\\
A: 97\\
\\
Now do the same for the following table and question.\\
\\
{\color{blue}{\{table\}}}\\
\\
Q: {\color{blue}{\{question\}}}
\end{prompt}
\caption{One-shot Table QA prompt for the TQA task.
\label{fig:one_shot_table_qa}
}
\end{figure}

\begin{figure}[h]
\begin{prompt}{One-shot Table+SQL QA prompt.}
\texttt{-{}-} Example \texttt{-{}-}\\
Answer the following question using the table below.\\
You may leverage an SQL tool.\\
The table is stored in a variable ‘sql\_table' and has the following schema:
\begin{verbatim}
Season | Team        | League_apps | Goals
1923   |Swindon Town | 55          | 3
1922   |Swindon Town | 14          | 4
\end{verbatim}
~~\\
Q: How many league appearances were there between 1907 and 1909 (inclusive)?\\
\begin{verbatim}
SQL: SELECT SUM(League_apps) FROM sql_table WHERE Season BETWEEN 1907 AND 1909 
\end{verbatim}

\begin{verbatim}
       | Result
result | 97
\end{verbatim}
~~\\
~~\\
Now do the same for the following table and question.\\
~~\\
{\color{blue}{\{table\}}}\\
~~\\
Q: {\color{blue}{\{question\}}}
\end{prompt}
\caption{One-shot Table+SQL QA prompt for the TQA task.
  \label{fig:one_shot_table_sql_qa}
} 
\end{figure}
\vspace{-4mm}

\begin{figure}[h]
\begin{prompt}{Generating a seed in TQA.}
Please generate an interesting statement about this table. The statement is a fact about one of the columns  in the following table.
\\
{\color{blue}{\{table\}}}\\
\\
An interesting statement as a result of this is:
\end{prompt}

   \caption{Prompt used to induce a pertinent and interesting seed topic in TQA. This is done zero-shot.
   \label{fig:tqa_inspiration}
   }

\begin{prompt}{Generating meaningful SQL in TQA.}
Please generate SQL statements for the following table:\\
\\
{\color{blue}{\{table\}}}\\
\\
Seed: {\color{blue}{\{seed\}}}\\
\\
An interesting SQL statement as a result of this is
\end{prompt}
   \caption{Prompt used to induce a meaningful SQL statement given the table and seed for the TQA task. This is done zero-shot.
    \label{fig:tqa_sql}
   }

\begin{prompt}{Generating a question  in TQA.}
I want to convert an SQL statement into a question.\\
Here is the original table:
\\
{\color{blue}{\{table\}}}\\
\\
SQL: {\color{blue}{\{SQL\}}}\\
\\
What is the question that this SQL statement would be the answer to?
\end{prompt}
\label{induce_question}
   \caption{Prompt used to induce a meaningful question using the table and generated SQL query for the TQA task. This is done zero-shot.}
\end{figure}

\begin{figure}[h]
\begin{prompt}{Three-shot CoT prompt used at evaluation time on MHQA.}
Answer the following multi-hop question `Q' by decomposing it into `Q1' and `Q2' and solving them step-by-step. Learn from the following 3 examples. As shown in the following example:\\
\\
\texttt{-{}-} Example \#1 \texttt{-{}-}\\
`Q' = `Who was the commander of the spaceflight that first landed humans on the Moon?' \\
\\
1. Splitting `Q' into `Q1' and `Q2':\\
‘Q1' : ‘What was the spaceflight that first landed humans on the Moon?';\\
‘Q2' : ‘Who was the commander of [A1]?';\\
\\
2. Answering Q1:\\
The answer `A1' to `Q1' : ‘What was the spaceflight that first landed humans on the Moon?' is ‘Apollo
11'. `A1' = ‘Apollo 11'\\
\\
3. Substituting A1 to Q2:\\
‘Q2' : ‘Who was the commander of Apollo 11?',\\
\\
4. Answers Q2:\\
The answer `A2' to Q2' : ‘Who was the commander of Apollo 11?' is ‘Neil Armstrong'.\\
`A2' = `A' = ‘Neil Armstrong'\\
\\
\texttt{-{}-} Example \#2 \texttt{-{}-}\\
`Q' = `What is the main ingredient in the flagship product of Ferrero?'\\
\\
1. Splitting `Q' into `Q1' and `Q2':\\
`Q1': ‘What is the flagship product of Ferrero?'\\
`Q2': ‘What is the main ingredient in [A1]?'\\
\\
2. Answering Q1:\\
The answer `A1' to `Q1' : ‘What is the flagship product of Ferrero?' is Nutella'.`A1' = Nutella'\\
\\
3. Substituting A1 to Q2:\\
‘Q2' : ‘What is the main ingredient in Nutella?',\\
\\
4. Answers Q2:\\
The answer `A2' to Q2' : ‘What is the main ingredient in Nutella?'.\\
`A2' = `A' = ‘Hazelnuts\\
\\
\texttt{-{}-}Example \#3 \texttt{-{}-}\\
\\
`Q' = `Who was the Roman Emperor when Jesus was born?'\\
1. Splitting `Q' into `Q1' and `Q2':\\
`Q1': ‘When was Jesus born? ‘\\
`Q2': ‘Who was the Roman Emperor in [A1]?'\\
\\
2. Answering Q1:\\
The answer `A1' to `Q1' : ‘When was Jesus born?' is 1 BCE. `A1' = 1 BCE\\
\\
3. Substituting A1 to Q2:\\
‘Q2' : ‘Who was the Roman Emperor in 1 BCE?',\\
\\
4. Answers Q2:\\
The answer `A2' to Q2' : ‘Who was the Roman Emperor in 1 BCE?'.\\
`A2` = `A` = ‘Caesar Augustus`\\
\\
You MUST apply this structure when asked to answer a multi-hop question `Q'.
Now answer the multi-hop question `Q` as shown in the examples above.\\
Q: {\color{blue}{\{question\}}}
\end{prompt}
\caption{\textit{Three-shot CoT prompt} used at evaluation time in MHQA.}
  \label{ft2_mhqa}
\end{figure}

\begin{figure}[h]
\begin{prompt}{Prompt used to merge Q1 and Q2 in MHQA.}
Merge `Q1` and `Q2' into a single multi-hop bridge question `Q'.\\
Learn from the following 3 examples. As shown in the following example:\\
\\
\texttt{-{}-} Example \#1 \texttt{-{}-}\\\\
‘Q1’ : "What was the spaceflight that first landed humans on the Moon?''\\
‘Q2': "Who was the commander of Apollo 11?''\\
\\
Solution:\\
1. Answer Q1; `A1' is "Apollo 11'' \\
2. If `A1' is in `Q2' print(A1); `A1' = Apollo 11 is in `Q2' so I print "Apollo 11''\\
3. Since you found `A1' in `Q2', rewrite `Q2' so that you delete `A1' and substitute `Q1' there;\\
Rewriting Q2. Original `Q2': "Who was the commander of Apollo 11?''. Since `A1' is in `Q2', I
delete it and write `Q1' there. Rewritten `Q2': "Who was the commander of the spaceflight
that first landed humans on the Moon?''\\
\\
The single multi-hop question is therefore the rewritten `Q2'.\\
`Q2` = `Q` = "Who was the commander of the spaceflight that first landed humans on the Moon?''\\
\\
\texttt{-{}-} Example \#2 \texttt{-{}-}\\\\
`Q1': What is the flagship product of Ferrero?\\
`Q2': What is the main ingredient in Nutella?\\
Solution:\\
1. Answer Q1; `A1' is "Nutella''\\
2. If `A1' is in `Q2' print(A1); `A1' = "Nutella'' is in `Q2' so I print "Nutella''\\
3. Since you found `A1' in `Q2', rewrite `Q2' so that you delete `A1' and substitute `Q1' there;\\
Rewriting Q2. Original `Q2': "What is the main ingredient in Nutella?''.\\
Since `A1' is in `Q2', I delete it and write `Q1' there.\\
Rewritten `Q2': "What is the main ingredient in the flagship product of Ferrero?''\\
\\
The single multi-hop question is therefore the rewritten `Q2'.
`Q2' = `Q' = "What is the main ingredient in the flagship product of Ferrero?''\\
\\
\texttt{-{}-} Example \#3 \texttt{-{}-}\\\\
`Q1': "When was Jesus born?''\\
`Q2': "Who was the Roman Emperor in 1 BCE?''\\
\\
Solution:\\
1. Answer Q1; `A1' is "1 BCE''\\
2. If `A1' is in `Q2' print(A1); `A1' = 1 BCE is in `Q2' so I print “1 BCE''\\
3. Since you found `A1' in `Q2', rewrite `Q2' so that you delete `A1' and substitute `Q1' there;\\
Rewriting Q2. Original `Q2': "Who was the Roman Emperor in 1 BCE?''. Since `A1' is in `Q2',
I delete it and write `Q1' there. Rewritten `Q2': "Who was the Roman Emperor when Jesus was
born?"\\
\\
The single multi-hop question is therefore the rewritten `Q2'.\\
`Q2' = `Q' = "Who was the Roman Emperor when Jesus was born?''\\
\\
\\
You MUST apply this structure when asked to merge `Q1' and `Q2'.\\
Now merge `Q1' and `Q2' into a single multi-hop bridge question `Q'.\\
`Q2' : {\color{blue}{\{question1\}}}\\
`Q2' : {\color{blue}{\{question2\}}}
\end{prompt}
  \caption{Prompt used to merge Q1 and Q2 in MHQA. 
  \label{fig:merge_mhqa}
  }
\end{figure}

\begin{figure}[h]
        
\begin{prompt}{Generating Q1 in MHQA.}
Identify one entity in the following text. Come up with a question so that the answer to this question is the entity chosen earlier. The question must be based on the following text. Write your results as 'Question:' and then the question and 'Entity:' and then the entity.\\\\
Text: {\color{blue}{\{document\_one\}}}
\end{prompt}

   \caption{Prompt used to generate $Q_1$. $Q_1$ is generated such that its answer $A1 = E$ where $E$ is the entity retrieved.
   \label{q1_gen_app}}

\end{figure}
\begin{figure}[h]

\begin{prompt}{Generating Q2 in MHQA.}
Come up with a question based on the following text that contains the word:
\\
{\color{blue}{\{entity\}}}\\
\\
Text: {\color{blue}{\{document\_two\}}}\\
\end{prompt}
   \caption{Prompt used to generate $Q_2$. $Q_2$ is generated such that its main topicis $E$ where $E$ is the entity retrieved. \label{q2_gen_app}}

\end{figure}

\end{document}